\documentclass[runningheads]{llncs}

\usepackage{eccv}

\usepackage{eccvabbrv}

\usepackage{graphicx}
\usepackage{booktabs}
\usepackage{tabularx}
\usepackage{algorithm}
\usepackage{booktabs} 
\usepackage{multirow}
\usepackage[T1]{fontenc} % 保证英文字体加粗有效
\usepackage{graphicx} 
\usepackage{adjustbox}
\usepackage{subcaption}
\usepackage{colortbl}  %彩色表格需要加载的宏包
\usepackage{xcolor}
\usepackage{array}   %对表列和表格线的设置需要用到array宏包
\usepackage{wrapfig}

\usepackage{algpseudocode}  
\usepackage{amsmath}  
\definecolor{red}{RGB}{255,0,0}
\definecolor{blue}{RGB}{0,0,255}
\definecolor{best}{HTML}{FDC9B4}
\definecolor{second}{HTML}{FEE2CE}
\definecolor{third}{HTML}{FFFAC2}

\newcommand{\RedTriangle}{{\color{red}$\triangleright$}}
\newcommand{\BlueTriangleRight}{{\color{blue}$\blacktriangleright$}}

\usepackage{hyperref}

\usepackage{orcidlink}

\begin{document}

\title{Improving Domain Generalization in Self-Supervised Monocular Depth Estimation via Stabilized Adversarial Training}
\titlerunning{SCAT}


\authorrunning{Yao et al.}
\author{Yuanqi Yao\inst{1}\orcidlink{0009-0005-3012-9395} \and
Gang Wu\inst{1}\orcidlink{0009-0007-5003-3117} \and
Kui Jiang\inst{1}\orcidlink{0000-0002-4055-7503} \and
Siao Liu\inst{2}\orcidlink{0000-0003-4285-3573} \and
Jian Kuai\inst{1}\orcidlink{0009-0008-8479-4579} \and
Xianming Liu\inst{1}\orcidlink{0000-0002-8857-1785} \and
Junjun Jiang\thanks{Corresponding author}\inst{1}\orcidlink{0000-0002-5694-505X}}


% \institute{Faculty of Computing, Harbin Institute of Technology, Harbin 150001, China 
% \email{yuanqiyao@stu.hit.edu.cn}} \and
% % \email{\{abc,lncs\}@uni-heidelberg.de}} \and
% \institute{Academy for Engineering & Technology, Fudan University
% \email{saliu20@fudan.edu.com}

\institute{Faculty of Computing, Harbin Institute of Technology, Harbin 150001, China
\email{\{yuanqiyao,kuaijian\}@stu.hit.edu.cn}\\
\email{\{gwu,jiangkui,csxm,jiangjunjun\}@hit.edu.cn} \and
Academy for Engineering \& Technology, Fudan University\\
\email{saliu20@fudan.edu.com}}

\maketitle
\authorrunning{First Author et al.}
\begin{abstract}

Learning a self-supervised Monocular Depth Estimation (MDE) model with great generalization remains significantly challenging. Despite the success of adversarial augmentation in the supervised learning generalization, naively incorporating it into self-supervised MDE models potentially causes over-regularization, suffering from severe performance degradation. In this paper, we conduct qualitative analysis and illuminate the main causes: (i) inherent sensitivity in the UNet-alike depth network and (ii) dual optimization conflict caused by over-regularization. To tackle these issues, we propose a general adversarial training framework, named Stabilized Conflict-optimization Adversarial Training (SCAT), integrating adversarial data augmentation into self-supervised MDE methods to achieve a balance between stability and generalization. Specifically, we devise an effective scaling depth network that tunes the coefficients of long skip connection and effectively stabilizes the training process. Then, we propose a conflict gradient surgery strategy, which progressively integrates the adversarial gradient and optimizes the model toward a conflict-free direction. Extensive experiments on five benchmarks demonstrate that SCAT can achieve state-of-the-art performance and significantly improve the generalization capability of existing self-supervised MDE methods.

\end{abstract}

\section{Introduction}
\label{sec:intro}

% Self-supervised 
Monocular depth estimation (MDE) plays an important role in various 3D perceptional fields such as robotic navigation~\cite{dong2022towards}, autonomous driving~\cite{wang2019pseudo} and 3D reconstruction~\cite{wu2018learning}.
% Inspired by , previous work usually formulate it as a XXX, which and achieve impressive performance. 
However, due to the dynamic nature of the real world, even minor perturbations in the environment can result in significant domain shifts in the visual observations, which makes the trained model hard to generalize into unseen scenarios and restricts its application in the physical world.

% Self-supervised monocular depth estimation is critical for deriving essential 3D structural information from scenes, serving as a cornerstone for numerous applications, including autonomous driving and robotic navigation~\cite{wang2019pseudo}-\cite{dong2022towards}. Despite its importance, the generalization of existing methods to novel domains remains a significant challenge, with a pronounced tendency for these methods to overfit to their training datasets.

% Self-supervised monocular depth estimation is critical for deriving essential 3D structural information from scenes, serving as a cornerstone for numerous applications including autonomous driving and robotic navigation~\cite{wang2019pseudo}-\cite{dong2022towards}.
% Despite its importance, the generalization of current methods to unseen scenarios remains a significant challenge.
% with a tendency to overfit to training datasets.
% with models often overfitting to their training datasets.

To improve the generalization capability, several studies~\cite{liu2021self,saunders2023self,liu2023iccv} utilize data augmentation methods to generate synthetic data and diversify the training environments, yielding considerable performance improvements. However, existing methods such as~\cite{saunders2023self} mostly select some specific data augmentation schemes for target scenarios, yielding poor generalization performance in environments varying far from the augmented images.
Compared with above mentioned offline data augmentation, adversarial data augmentation (ADA) does not make any target distribution assumption and synchronously optimizes the augmenter during the training phase, providing a promising pre-processing solution. 
Unfortunately, there is a dilemma in naively incorporating ADA into self-supervised MDE. Although adversarial data augmentation can effectively improve generalization capability in multiple supervised visual tasks~\cite{saunders2023self}, self-supervised MDE algorithms are quite sensitive to such excessive perturbation, resulting in significant performance degradation and training collapse.
% (as illustrated in Figure~\ref{fig:fig11}). 
Therefore, it is necessary to rethink why self-supervised MDE cannot benefit from ADA as much as supervised learning.

% Recent studies~\cite{liu2021self,saunders2023self} have demonstrated that while offline data augmentation targeting common corruptions can enhance model generalization in specific scenarios, excessive augmentation may actually diminish sample efficiency and potentially cause divergence. 
% Although such augmentations hold promise for bolstering generalization, the resultant variability in data complicates optimization, introducing a risk of instability. In contrast to supervised learning, achieving a balance between stability and generalization in self-supervised learning necessitates extensive experiments.

\begin{figure}[t]
\centering
\includegraphics[width=0.9\columnwidth, keepaspectratio, interpolate=true]{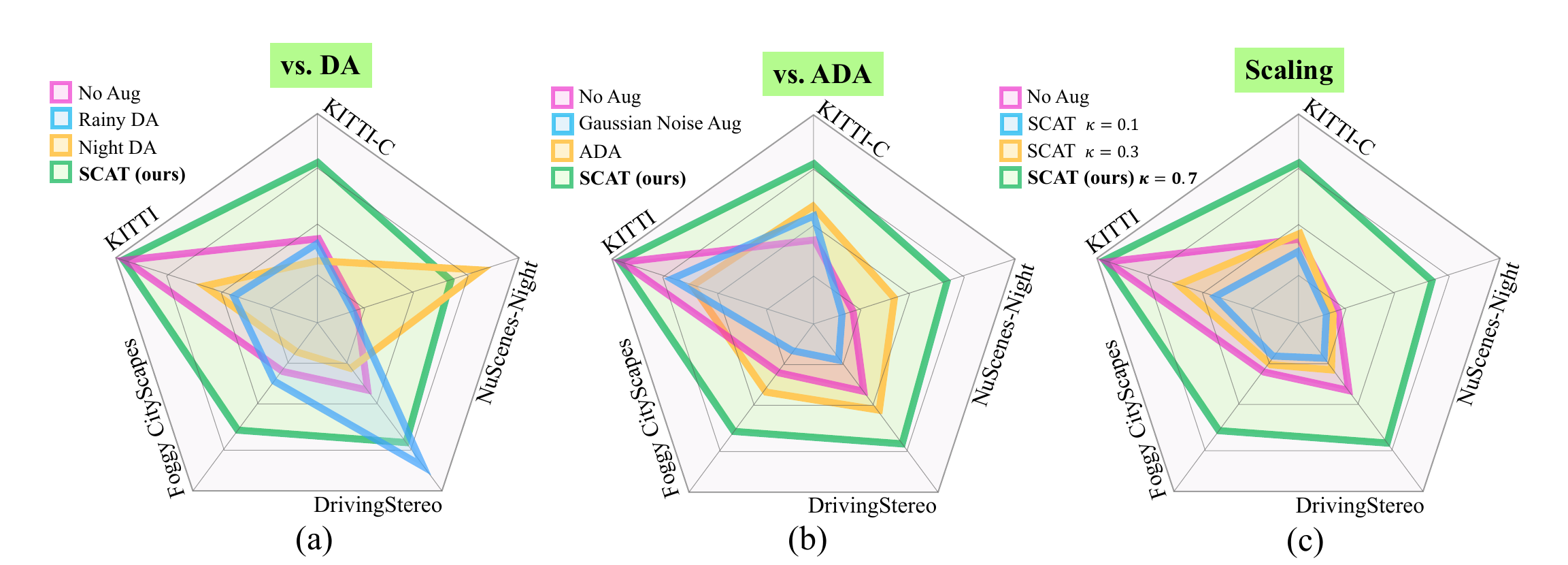}
\caption{\textbf{Visualization of the domain generalization over various methods.}(a) Comparisons with offline scenario-specific data augmentation methods.(b) Comparisons  with vanilla Gaussian noise and vanilla adversarial data augmentation.(c) Comparisons of different LSCs scaling factors. The results show that our SCAT has excellent generalization performance under multiple unseen domains and retains the performance on the original training set.}\vspace{-20pt}
% As the SOTA self-supervised depth estimation model, M& MonoVit excels on the KITTI dataset, but it struggles to accurately infer depth information from various types of damaged images in out-of-distribution (OoD) scenarios.

\label{fig:fiveline}
\end{figure}
In this work, we first conduct extensive quantitative analysis to investigate the causes of performance degradation when applying adversarial data augmentation to common self-supervised MDE models. There are two primary factors for this phenomenon: (i) inherent sensitivity of long skip connections (LSC) in UNet-alike depth estimation networks; (ii) dual optimization conflict caused by over-regularization. Specifically, as a core component in the MDE, LSC is widely adopted in the UNet-alike Depth Networks to combine multi-scale features and preserve low-level detail, yielding better prediction performance. However, the presence of these shortcut connections amplifies adversarial gradients, when coupled with pixel-level adversarial data augmentation, leading to severe training instability and collapse. Moreover, compared to offline data augmentation, adversarial augmentation data commonly act as the worst-case training examples~\cite{rice2021robustness,wang2022evaluating} and often provides over-regularization for the model training, which results in optimization gradients that oppose those of the original data, leading to severe gradient conflicts and a subsequent decline in convergence performance.

% , as a worst-case example for the model training, often provides over-regularization for the model training and results in optimization gradients that oppose those of the original data, leading to severe gradient conflicts and a subsequent decline in model performance.
% fuse features of various granularities, improving depth map accuracy by integrating context and detail. and identify the main factors contributing significantly to the instability seen in prior applications of data augmentation: 
% In this study, we explore the theoretical underpinnings behind the instability observed in both adversarial and offline data augmentation techniques when applied to self-supervised MDE tasks. 
% Drawing from our analysis, we propose a general adversarial training framework to simultaneously improve the model's generalization capabilities and stability. 
% We identify three primary factors contributing significantly to the instability seen in prior applications of data augmentation: (i) constantly changing augmented data resulting in high variance reconstruction targets; and (ii) sensitivity of UNet-based depth networks to perturbed inputs; (ii) dual optimization conflict caused by over-regularization. 

To tackle these issues, we propose a general adversarial training framework, named \textbf{S}tabilized \textbf{C}onflict-optimization \textbf{A}dversarial \textbf{T}raining (\textbf{SCAT}), tailored for self-supervised monocular depth estimation. SCAT is designed to enhance both generalization capability and stability.  In particular, we develop an innovative scaling depth network that adjusts the coefficients of long skip connections within UNet architecture, thereby ensuring a more stable training regime through a theoretically supported. Furthermore, we advance a Conflict Gradient Surgery method, progressively blending the adversarial gradient to guide the model optimization along a direction devoid of conflicts.
% Our approach is distinguished by two key innovations: Firstly, SCAT focuses on introducing ADA into self-supervised MDE methods, however, as reported in~\cite{rice2021robustness, wang2022evaluating},
% % SCAT focuses on reconstructing the original input images rather than those altered by adversarial augmentation, thereby sidestepping the pitfalls of high-variance reconstruction targets that can misguide the training process. Secondly, based on~\cite{rice2021robustness, wang2022evaluating}, 
% although basic adversarial augmentation can enhance model generalization, the worst-case induced by adversarial methods also leads to dual optimization conflict. These conflicts in gradient components prevent the model from achieving stable convergence. Therefore, we propose Conflict Gradient Surgery (CGS) to refine the adversarial training gradient update process. CGS incrementally applies adversarial augmenters from multiple iterations, enabling a balanced achievement in model generalization and stability. 
It is worth noting that one of the merits of our method SCAT is model-agnostic, and the stabilized adversarial training can be applied to diverse SOTA methods, yielding consistent generalization enhancement.
To validate the effectiveness of SCAT, we conduct extensive experiments on both KITTI and KITTI-C datasets, demonstrating its cross-domain generalization capability on Foggy CityScapes, DrivingStereo and NuScenes datasets. In summary, our contributions are summarized as follows:
% \begin{itemize}
%     \item We develop an adversarial training framework for self-supervised depth estimation, significantly improving model robustness without any additional data augmentation.
%     \item We introduce a generalization-enhanced adversarial training strategy, allowing ARDepth to perform well on corrupted datasets while maintaining its performance on clean KITTI dataset.
%     \item Our ARDepth achieves SOTA performance on both KITTI and KITTI-C datasets, and demonstrates cross-domain generalization capability on Foggy CityScapes and NuScenes datasets.
% \end{itemize}
\begin{itemize}
    % \item[$\bullet$] We point out the high-variance targets induced by data augmentation and provide a theoretical analysis of the model instability due to naive offline or adversarial data augmentation in self-supervised MDE.
    \item[$\bullet$] We point out the inherent sensitivity of UNet-based depth networks and the dual optimization conflict caused by over-regularization, providing a theoretical analysis of the self-supervised MDE model's instability.
    \item[$\bullet$] We propose a general adversarial training framework named Stabilized Conflict-optimization Adversarial Training (SCAT), which efficiently incorporates adversarial augmentation into self-supervised MDE, and significantly improves the model generalization capabilities across multiple unseen domains.
    \item[$\bullet$] We developed Conflict Gradient Surgery (CGS) to address the dual optimization conflicts induced by adversarial over-regularization, achieving a balance between model stability and generalization.
   % \item[$\bullet$] Compared to previous state-of-the-art methods, SCAT has demonstrated highly competitive generalization performance across multiple cross-domain datasets and has significantly improved training efficiency.
\end{itemize}

% \section{Initial Submission}
\section{Related Works}

% \subsection{Supervised Depth Estimation}

% The advent of autonomous driving and robotics has sparked significant interest in depth estimation technology. %Supervised depth estimation methods %involve utilizing 
% %require actual depth values as labels to establish pixel-depth relationships. 
% For example, Eigen \emph{et al.}~\cite{eigen2014depth} pioneered this field by introducing convolutional neural networks (CNNs) for monocular depth estimation and later accelerated convergence by leveraging pretrained networks. To %address issues like 
% promote object boundary distortion and blurry reconstructions, Hu \emph{et al.}~\cite{hu2019revisiting} introduced a multi-scale feature fusion network.
% Although the supervision-based methods have achieved significant progress for depth estimation, %However, supervised monocular depth estimation 
% these approaches require costly real-depth labels, greatly limited to specific scenes. %Consequently, attention has shifted toward self-supervised methods.

\subsection{Self-Supervised Monocular Depth Estimation}
% The advent of autonomous driving and robotics has sparked significant interest in depth estimation technology.
% %While supervised methods have yielded impressive results, obtaining accurate ground truth depth during training remains challenging.  To overcome this, 
% Since obtaining accurate ground truth depth during training is a time-consuming and troublesome task, researchers have explored self-supervised deep learning for monocular depth estimation. These methods~\cite{garg2016unsupervised,luo2018single,zhou2017unsupervised,godard2019monodepth2} %reframe 
% regard depth estimation as alternative visual tasks like image reconstruction, typically relying on information aggregation from multiple images.
Self-supervised monocular depth estimation (MDE) has made remarkable progress in recent years, enabling depth learning from unlabeled data. The seminal work by Zhou et al.~\cite{zhou2017unsupervised} laid the foundation by jointly optimizing depth and pose networks using an image reconstruction loss. Since then, two main approaches have emerged: stereo training and monocular training.
Stereo training methods utilize synchronized stereo image pairs to predict disparity maps, with notable contributions including photometric consistency loss~\cite{garg2016unsupervised}, left-right consistency~\cite{godard2017unsupervised}, and continuous disparity prediction~\cite{garg2020wasserstein}. Monocular training, on the other hand, relies on the consistency between synthesized and actual scene views, with SfMLearner~\cite{zhou2017unsupervised} pioneering the joint training of a DepthNet and a PoseNet using a photometric loss.
Researchers have also explored complementary techniques to further boost performance, such as occlusion modeling~\cite{godard2019digging,poggi2018learning,ramamonjisoa2020predicting,wong2019bilateral}, scale ambiguity~\cite{zhang2022towards} advanced network architectures~\cite{han2022brnet,pillai2019superdepth,zhou2022self}, semantic information integration~\cite{chen2019towards,jung2021fine,zhu2020edge,ma2022towards,chen2023self}, and object size cues~\cite{zhu2020edge,peng2021excavating}. The proposed use of orthogonal planes for improved depth representation in driving scenarios contributes to the ongoing advancements in this field.
% Among these, binocular images or consecutive video frames are commonly used inputs.
% \subsubsection{Methods Based on Stereo Cameras}
% Stereo vision, based on disparity, captures the scene with two cameras from different angles, enabling 3D information acquisition.  Garg~\emph{et al.}~\cite{garg2016unsupervised}  transformed monocular depth estimation into monocular disparity estimation, using stereo images and predicted disparity maps for training.  Luo~\emph{et al.}~\cite{luo2018single} viewed depth estimation as stereo matching, dividing it into image synthesis and stereo matching sub-tasks, offering a novel self-supervised approach.  However, these methods still %exhibit 
% suffer from %issues like 
% occlusions and blurry image boundary problems.
% \subsubsection{Methods Based on Multi-Frames}
% Self-supervised  Monocular Depth Estimation (MDE) methods based on multiple frames transform depth estimation into the image reconstruction task between consecutive frames, typically using monocular image sequences (videos) as training data.  %Pose estimation modules are incorporated, usually employing deep learning-based pose networks. 

\subsection{Domain Generalization in Monocular Depth Estimation}
% Generative Adversarial Networks emerging with powerful representation capability, have been applied to depth estimation task.  Aleotti~\emph{et al.}~\cite{aleotti2018generative}  introduced a self-supervised monocular depth estimation framework with a discriminator network. The adversarial training is introduced to promote the robustness of generator.  Cheng~\emph{et al.}~\cite{cheng2023adversarial} proposed to degrade the input 2D images to generate the adversarial samples, enabling the network to possess discriminative ability and robustness against real-world attacks.
%proposed an adversarial training approach using view synthesis to perturb 2D images, enhancing robustness against real-world attacks.
A line of studies investigated how to utilize data augmentation or synthetic data generation to improve the domain generalization capability of MDE~\cite{liu2021self,atapour2018real,gurram2021monocular,spencer2020defeat,vankadari2023sun,wang2021regularizing,zhao2022unsupervised,saunders2023self}. 
ADDS-DepthNet~\cite{liu2021self} generates night image pairs from day images using GANs, training the network to be effective in both day and night domains.  It uses day depth estimates as pseudo-supervision for night scenes, which limits the night depth estimation to the accuracy of the day estimates. Additionally, ADDS-DepthNet focuses on reconstructing GAN-generated night images, which generates reconstruction targets with high variance, resulting detrimental to the self-supervised MDE algorithm.
ITDFA~\cite{zhao2022unsupervised} employs a fixed depth decoder and adapts the encoders for each domain to enforce feature consistency across domains, which is limited to the performance.
% EPC-depth~\cite{peng2021excavating}  demonstrated improved stereo depth performance using data grafting for input images.
Robust-Depth~\cite{saunders2023self} put forward an offline data augmentation method. By exploiting the correspondence between unaugmented and augmented data they introduce a pseudo-supervised loss for both depth and pose estimation.  However, previous approaches limited the ability to handle environments with pre-defined variations and require substantial architectural modifications to obtain realistic depth estimates, leading to severe domain bias when applied to real-world scenes~\cite{liu2021self,saunders2023self}.
\subsection{Adversarial Training}
Adversarial training has emerged as a popular technique to improve the generalization capability of deep neural networks. This approach involves incorporating both benign and adversarial examples during the training process, enabling models to learn more robust features~\cite{carlini2017towards}. Researchers have successfully applied adversarial training to various computer vision tasks, such as image classification~\cite{rusak2020simple,carlini2017towards}, object detection~\cite{zhang2019towards,chen2021robust}, and segmentation~\cite{xu2021dynamic,hung2018adversarial}.
However, generating adversarial examples typically requires ground truth labels, which limits the applicability of adversarial training to supervised learning settings. To address this limitation, semi-supervised adversarial learning methods~\cite{carmon2019unlabeled,alayrac2019labels} have been proposed, which leverage a small portion of labeled data to enhance robustness.   Additionally, contrastive learning~\cite{ho2020contrastive,kim2020adversarial} has been combined with adversarial examples to improve self-supervised learning and model robustness.
In the context of MDE, adversarial training remains largely unexplored, particularly in self-supervised MDE which absent the ground truth depth information. 

% \subsection{Domain Generalization}

\section{Preliminaries}
\subsection{Problem Formulation}
Following~\cite{zhou2017sfm,godard2019monodepth2}, we regard the self-supervised depth estimation as an image reconstruction task, where the objective is to reconstruct the target current frame image $I_t$ from consecutive frames $I_t' \in \{I_{t-1}, I_{t+1}\}$. 
Specifically, we construct a dual-branch framework, involving a DepthNet and a PoseNet. 
The former is used for estimating depth map $D_t$ while the latter is introduced to predict the relative poses $T_{t' \rightarrow t}$ through the inverse warping as below, the predicted depth map $D_t$ in conjunction with the relative poses $T_{t' \rightarrow t}$ can synthesize the target image $I_{t' \rightarrow t}$, depicted as:
\begin{equation}
\label{eq1}
I_{t^{\prime}\rightarrow t}\;=\;I_{t^{\prime}}\Big\langle proj\big(D_{t},T_{t\rightarrow t^{\prime}},K)\Big\rangle,
\end{equation}
where $proj(\cdot)$ generates the resulting 2D coordinates of the projected depths $Dt$ in $It'$ and $\langle \rangle$ refers to the sampling operator.We minimize the photometric reprojection error $\mathcal{L}_p$ to supervise the DepthNet, 
\begin{equation}
\label{eq3}
\mathcal{L}_{p}\;=\;\sum_{t^{\prime}}pe(I_{t},I_{t^{\prime}\to t}),
\end{equation}
where $pe$ is the photometric reconstruction error. As suggestion in~\cite{zhao2016loss,godard2017unsupervised}, we adopt the weighted combination of $L_1$ loss and SSIM~\cite{wang2004image} to formulate the photometric error function $pe$, denoted as:
\begin{equation}
\label{eq4}
p e(I_{a},I_{b})=\frac{\alpha}{2}(1-\mathrm{SSIM}(I_{a},I_{b}))+(1-\alpha)||I_{a}-I_{b}||_1.
\end{equation}

\subsection{Analysis on Unstable Self-Supervised MDE}
In this section, we aim to investigate the primary causes why naively applying adversarial data augmentation in self-supervised MDE leads to instability.

\noindent\textbf{{Pitfall1: Sensitive UNet-alike Depth Network.}\label{UNet}}
In previous self-supervised MDE methods~\cite{godard2019monodepth2,yan2021cadepth,lyu2021hrdepth}, UNet~\cite{ronneberger2015u} is the most popular depth network backbone, since its long skip connects (LSCs) to connect distant network blocks can fuse multi-granularity semantic information and spatial information:
\begin{equation}
\label{eq13}
\mathrm{DepthNet}(x)=f_{0}(x),\ f_{i}(x)=b_{i+1}\circ[ a_{i+1}\circ x+f_{i+1}\,(a_{i+1}\circ x)]\,,
\end{equation}
where $x \in R^m$ denotes the input, $a_i$ and $b_i (i \geq 1)$ are the trainable parameter of the $i$-th block.
% $k_i > 0 (i \geq 1)$ are the scaling coefficients and are set to 1 in standard UNet. $f_N$ is the middle block of UNet. 
For the vector operation $\circ$, it can be designed to implement different networks.
% To solve this issue, we theoretically show that the coefficients of LSCs in depth network have big effects on the stableness of the gradient propagation and robustness of depth network.  
% As discussed in Section~\ref{Target} , self-supervised MDE with data augmentations aims to reprojection $I_{t^{\prime}\rightarrow t}$ from the noisy input $\tilde{I}_{t'}$. 
Assume an extra noise $\epsilon_\delta \sim N (0, \;\sigma^2_\delta I)$ is injected into $I_t$ which yields a new input $\tilde{I}_t'$, self-supervised MDE with data augmentations aims to reprojection $I_{t^{\prime}\rightarrow t}$ from the noisy input $\tilde{I}_{t'}$. Accordingly, if the variance $\sigma^2_\delta$ of extra noise is large, it can hinder re-projecting the desired target image $I_{t' \rightarrow t}$. In adversarial training, the variance of this extra noise varies along training iterations, further exacerbating the instability of self-supervised MDE training. Recent work~\cite{huang2024scalelong} also reveals a similar phenomenon, which demonstrates that the UNet-based segmentation architecture is sensitive to perturbed noise, suffering from catastrophic training collapse. It is noteworthy that, although our theoretical analysis is focused on UNet-based depth estimation architectures, the LSC  component is extensively employed across various mainstream architectures~\cite{zhao2021m& MonoVit}, thereby lending a degree of extensibility to our empirical results.
%-siao- 这里有一种写limitation的感觉，和你的方法没有甚么关系，续写上面那句话，直观上有啥问题
% Moreover, as there are many variants of depth networks, e.g.~\cite{zhao2021m& MonoVit}, it is hard to provide a unified analysis for all of them. Thus, here we only analyze the most widely used UNet, and leave the analysis of other UNet variants in future work.

% However, theoretical understandings of the instability of depth network in self-supervised MDE models and also the performance improvement of LSC scaling remain absent yet. 

% Inspired by~\cite{huang2024scalelong}, UNet-based depth network often suffers from the unstable training issue in data augmentations, 
% while scaling the long skip connection (LSC) coefficients $k_i(\forall i)$ in UNet can help to stabilize the training. However, theoretical understandings of the instability of Depth network
% and also the effects of scaling LSC coeffeicents 
% remain absent yet, 
% hindering the development of new and more advanced depth network in a principle way. 
% To address this problem, in this section, we will theoretically and comprehensively analyze
% 1) why depth network is unstable and also 2) why scaling the coeffeicents of LSC in UNet helps stablize training. 
% why depth network is unstable.
% Here we use a practical example to show the importance of the robustness of UNet in DMs. 

\noindent\textbf{Pitfall2: Dual Optimization Conflict by Over-Regularization.}
%-siao- 这里一坨
Compared to the offline data augmentation in Robust-Depth~\cite{saunders2023self}, Adversarial Data Augmentation (ADA) makes no assumptions about the target distribution and synchronously optimizes augmenters during the training phase, offering a promising pre-processing solution. Unfortunately, naively incorporating ADA into self-supervised Monocular Depth Estimation (MDE) presents a dilemma. While adversarial data augmentation can effectively enhance generalization in various supervised visual tasks, a dual optimization conflict arises due to the opposing gradient optimization directions between the adversarial-perturbed data and the original data. Similar to~\cite{yu2020gradient}, we also define $\phi_{ij}$ as the angle between origin data $g_i$ and adversarial-perturbed data gradient $g_j$, the gradients as conflicting when $cos \phi_{ij} \textless 0$. Therefore, for self-supervised MDE, over-regularization leads to ADA dominating the gradient updates, resulting in significant performance degradation and training collapse. We refer to this issue as a dual optimization conflict caused by over-regularization.
% In the following section, we address these pitfalls.

\section{Method}
% In this section, we propose a general adversarial training framework named Stabilized Conflict-optimization Adversarial Training (SCAT), which is compatible with any self-supervised MDE method.
% and without modifying their underlying neural network architecture.

\begin{figure*}[t]
\centering
\includegraphics[width=\textwidth, keepaspectratio, interpolate=true]{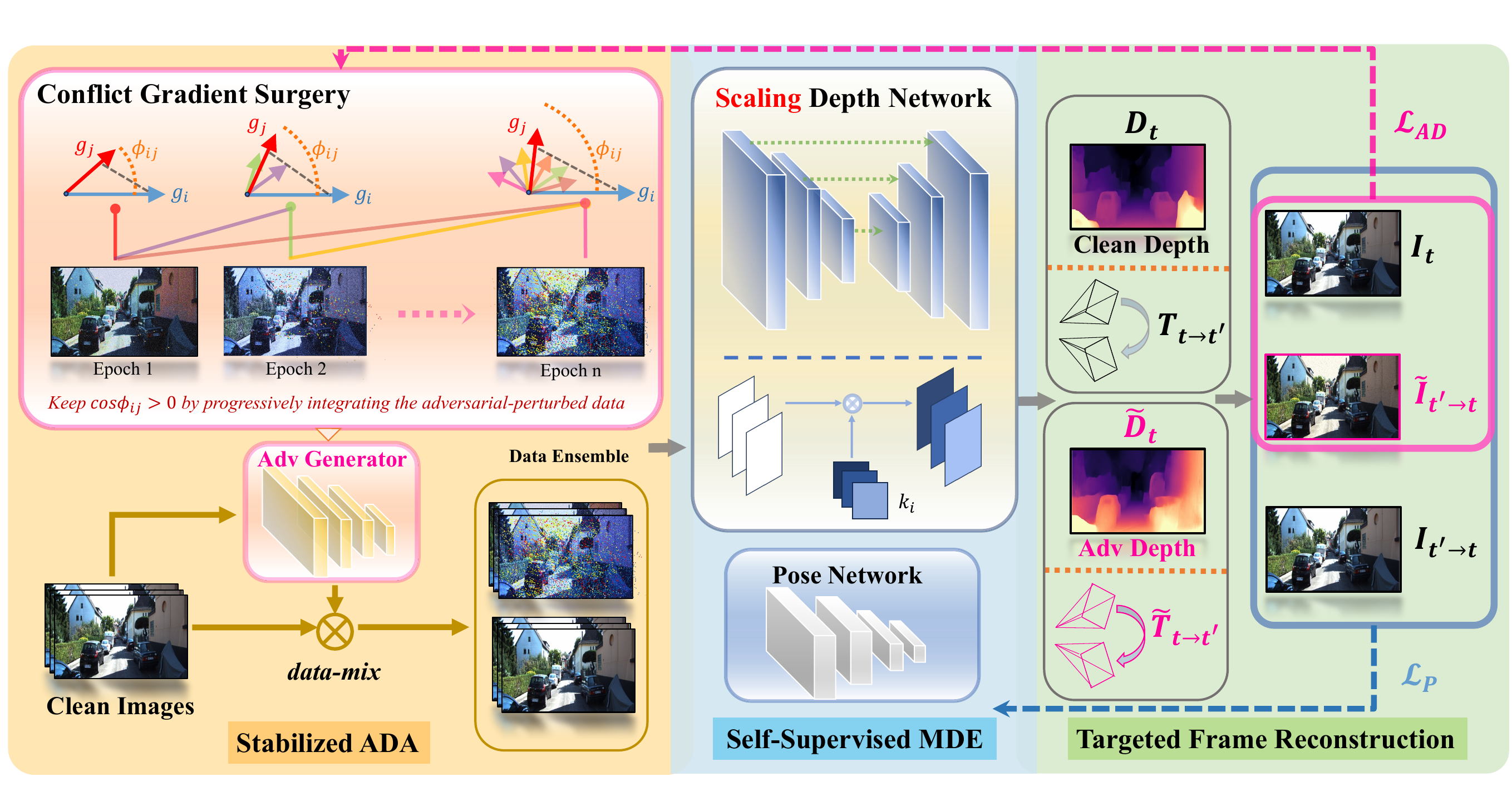}
\caption{\textbf{Overview of our SCAT architecture.}
\vspace{5pt}
% Our ARDepth consists of two parts. In part1, we introduce adversarial perturbations into the input multi-frame images using an adversarial noise generator. These perturbed adversarial images are then blended with the original clean images using our generalization-enhanced adversarial training strategy, serving as the training input for pose network depth network in part 2.
% Within Part 2, we leverage relative pose $T_{t \rightarrow t'}, T_{t \rightarrow t'_{ad}}$ and depth map $D_t, D_{t_{ad}}$ to reconstruct adversarial-reconstructed images and clean-reconstructed images for the current frame $t$.  We compute reprojection losses relative to the original input image for the current frame, yielding the corresponding $\mathcal{L}_{AD}$ and $\mathcal{L}_p$.  The $\mathcal{L}_{AD}$ loss acts as a self-supervised constraint for our adversarial noise generator, while $\mathcal{L}_p$ imposes constraints on depth networks.}%%%%\vspace*{-4mm}
Our method introduces an adversarial noise generator, which is optimized through $\mathcal{L}_{AD}$ and acts as an adversarial constraint, while $\mathcal{L}_p$ imposes self-supervised constraints to optimize self-supervised MDE model. CGS incrementally applies adversarial augmenters from multiple iterations, enabling a balanced achievement in model generalization and stability. Meanwhile, we utilize a scaling depth network(SDN) to stabilize the training process.}%%%%\vspace*{-4mm}
\label{fig:fig2}
\vspace{-5pt}
\end{figure*}

\subsection{Architectural Overview}\label{JointAdversarialTraining}
In this section, we propose a general adversarial training framework named Stabilized Conflict-optimization Adversarial Training (SCAT), which is compatible with any self-supervised MDE method.
An overview of the SCAT architecture is provided in Fig.~\ref{fig:fig2}.
To circumvent erroneous bootstrapping from adversarial augmented data, augmentation was applied to improve generalization by reconstructing the unaugmented $I_t$, rather than the augmented image $\tilde{I}_t$.
% , which addresses Pitfall 1.
\begin{equation}
\label{eq6}
\tilde{I}_{t'} = I_{t'} + \delta
\end{equation}
\begin{equation}
\label{eq11}
\tilde{I}_{t^{\prime}\rightarrow t}\;=\;I_{t^{\prime}}\Big\langle proj\big(\tilde{D}_{t},\tilde{T}_{t\rightarrow t^{\prime}},K)\Big\rangle,
\end{equation}
\begin{equation}
\label{eq_p}
\mathcal{L}_{p}\;=\;\sum_{t^{\prime}}pe(I_{t},I_{t^{\prime}\to t}) + pe(I_{t},\tilde{I}_{t^{\prime}\to t}).
\end{equation}
If $\delta$ is learned (i.e. SCAT is implemented with an adversarial generator $\delta = g_{\phi}(z)$, to introduce adversarial training into self-supervised MDE, we calculate the reprojection loss for the current frame image reconstructed from the perturbed images and use it as the self-supervised constraint loss $\mathcal{L}_{AD}$:
\begin{equation}
\label{eq_ad}
\mathcal{L}_{AD}\;=\;\sum_{t^{\prime}} pe(I_{t},\tilde{I}_{t^{\prime}\to t}).
\end{equation}
Our goal is to find a noise distribution $p_{\phi}$($\delta$), $\delta$ $\in$  $R^N$ and $\Vert \delta \Vert_2$=$\epsilon$, such that noise samples added to input clean image $I_t$ maximally confuse the depth network. More concisely, we optimize the following equation, depicted as 
\begin{equation}
\label{eq5}
\min _{\theta} \max _{\phi} \mathbb{E}_{\tilde{I}_{t'}, I_t} \mathbb{E}_{\delta \sim p_\phi(\delta)}\left[\mathcal{L}\left(f_{\theta}(\tilde{I}_{t'}),I_t\right)\right].
\end{equation}
% Maximizing $\mathcal{L}_{AD}$ translates to maximizing depth estimation error when the model processes perturbed images.  This ensures that adversarial perturbations could induce as many errors as possible in the depth estimation, achieving an adversarial effect with the depth estimation model.
% The outer layer minimizes the depth network's loss while keeping the perturbation fixed, making the depth network robust and adaptable to such perturbations. 
Specifically, ${max}_\phi$ represents the adversarial noise generator's attempt to maximize the $\mathcal{L}_{AD}$, thereby generating adversarial samples that maximally perturb the depth network. Conversely, $min_\theta$ denotes the depth network's effort to minimize the $\mathcal{L}_{p}$ on these adversarial samples, aiming to achieve robustness in depth estimation across various adversarial inputs. 
Through training iterations provided in Algorithm~\ref{alg:scat} in our SCAT framework, self-supervised MDE models becomes proficient at accurately inferring various perturbed images.
% , which improves the capability of generalization.
% , enhancing self-supervised depth estimation robustness. 

However, as the variance of this extra adversarial noise varies along training, the standard UNet-based depth network is too sensitive to these perturbations. Hence, we propose a Scaling Depth Network (SDN), which appropriately scales the coefficients ${\kappa_{i}}$ to lower reconstruction error.
% ensure that the model has enough representational power while preventing too big a gradient and avoid unstable parameter updates. 
In addition, 
% as in Pitfall 2, 
conflict arises by opposing gradient optimization directions between the adversarial-perturbed data and original data, further exacerbating the training instability. 
% To mitigate the instability caused by the model's sensitivity to perturbations,
% we firstly propose a Scaling Adversarial Augmentation method (S2A), which appropriately scales the coefficients ${k_i}$ to ensure that the model has enough representational power while preventing too big a gradient and avoid unstable parameter updates. Further, although S2A can enhance model generalization, the worst-case induced by adversarial methods also leads to dual optimization conflict. 
% These conflicts in gradient components prevent the model from achieving stable convergence. Drawing on this, 
To ensure the model has enough representational power while preventing unstable parameter updates, we propose Conflict Gradient Surgery (CGS) to refine the adversarial training gradient update process, which we introduce in the following section.

\subsection{Scaling Depth Network}\label{SDN}
To alleviate the impact of extra noise, we devise scaling depth network (SDN) to scale the coefficients ${\kappa_{i}}$ of LSCs :
% As illustrated in Fig.~\ref{fig:fig11}, it shows that extra noise $\epsilon_\delta$ can significantly affect the performance of UNet-based depth network, and the scaled LSCs can alleviate the impact of extra noise to some extent.
\begin{equation}
\label{eq20}
\mathrm{DepthNet}(x)=f_{0}(x),\ f_{i}(x)=b_{i+1}\circ[k_{i+1} \cdot a_{i+1}\circ x+f_{i+1}\,(a_{i+1}\circ x)]\,,
\end{equation}
where $\kappa_{i} \textgreater 0, (i \geq 1)$ are the scaling coefficients and are set to 1 in standard UNet-based depth network. For scaling depth network in Eq.\ref{eq20}, assume $M_{0}\,=\,\mathrm{max}\{||b_{i}\circ a_{i}||_{2},1\leq\,i\leq N\}$ and $f_N$ is $L_0-Lipschitz \; continuous$, $c_0$ is a constant related to $M_0$ and $L_0$. Suppose $I_{t}^{\epsilon_{\delta}}$ is an perturbed input of the vanilla input $I_t$ with a small perturbation $\epsilon_{\delta} = ||I_t^{\epsilon_{\delta}} - I_t||_{2}$. Then we have
\begin{equation}
\label{eq24}
||\mathrm{f_{\theta}}({\bf I}_{t}^{\epsilon _ {\delta}})-\mathrm{f_{\theta}}({\bf I}_{t})||_{2}\;\le\;\epsilon _ {\delta}\left[\sum_{i=1}^{N}\kappa_{i}M_{0}^{i}+c_{0}\right],
\end{equation}
where $N$ is the number of the long skip connections. See the proof in Appendix.

Eq.\ref{eq24} shows that for a perturbation magnitude $\epsilon_\delta$, the reconstruction error bound of depth network is ${{\mathcal{O}}(\epsilon_{\delta}\bigl(\sum_{k=1}^{}K_{i}M_{0}^{i}\bigr)}$. For standard UNet-based depth network $(\kappa_{i}=1\forall i)$, this bound becomes a very large bound $\mathcal{O}(N M_0^N )$. 
This implies that a standard UNet-based depth network is sensitive to extra noise in the input, especially when LSC number N is large. 
% Hence, it is necessary to control the coefficients $k_i$ of LSCs which accords with the observations in Fig.~\ref{}. 
Intuitively, setting appropriate scaling coefficients ${\kappa_{i}}$ 
in our scaling depth network 
can enhance the robustness of the model to input perturbations, thereby improving better stability for self-supervised MDE. 
To derive the scaling coefficient $\kappa$, we adopt a heuristic algorithm to assign a uniform constant value of $\kappa_*$ across all LSCs, with the default set to 0.7. In terms of implementation, a line of parameter optimization techniques~\cite{gridsearch,huang2024scalelong} can be applied, e.g. incorporating an adaptive projection layer~\cite{huang2024scalelong} for each LSC component. The empirical discussions in the Appendix demonstrate that a singular scaling parameter $\kappa$ is sufficient to strike to balance performance and efficiency for the self-supervised MDE task.
\subsection{Conflict Gradient Surgery}\label{CGS}
\begin{figure}[h]
\vspace*{-6mm}
  \centering
  % \fbox{\rule{0pt}{0.5in} \rule{0.9\linewidth}{0pt}}
  \includegraphics[width=0.9\columnwidth, keepaspectratio, interpolate=true]{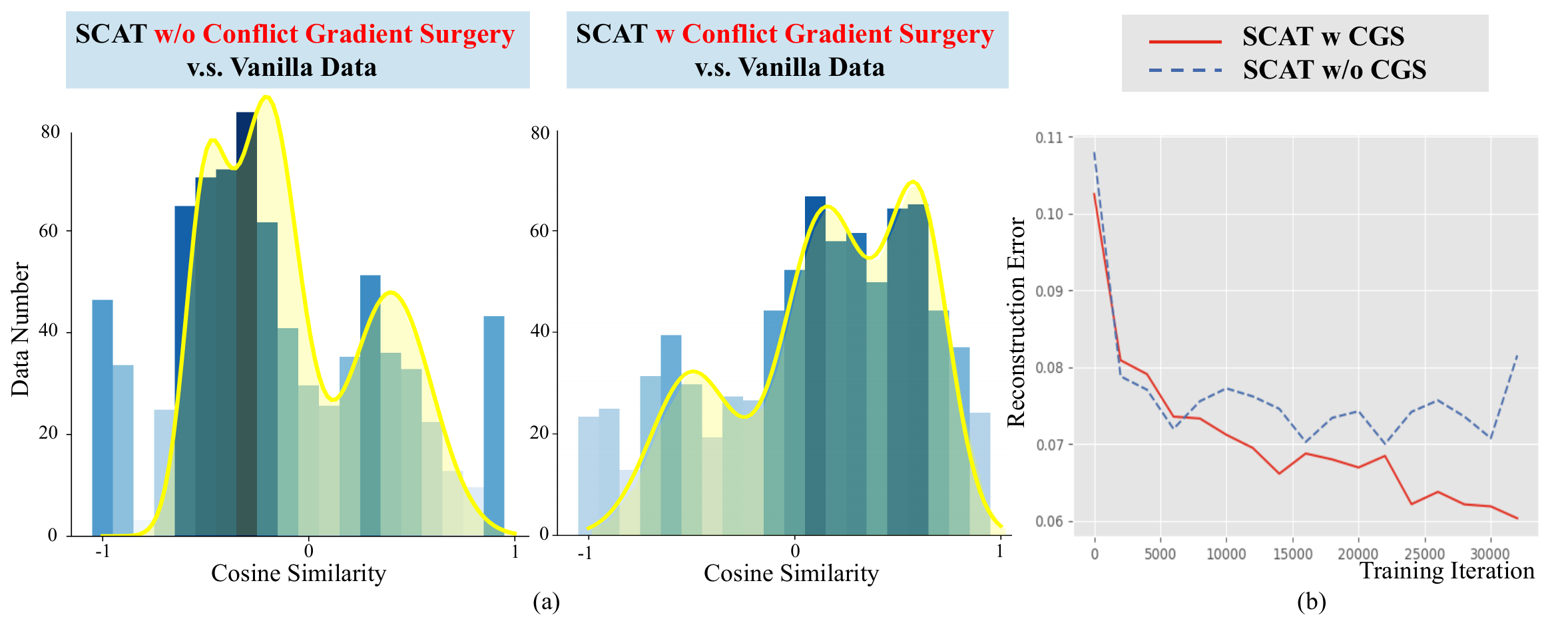}
  \vspace*{-3mm}
   \caption{\textbf{(a) Statistics of gradient cosine similarity.} Through CGS, we have shifted the distribution of cosine similarities from being negatively skewed to positively skewed, mitigating the previously prevalent issue of adversely oriented gradients.
   \textbf{(b) Illustration of training oscillation issue arising from Dual Optimization Conflict.}}
   \label{fig:grad_dist}
   \vspace*{-6mm}
\end{figure}

Here, we propose to modify the standard adversarial data augmentation (ADA) by incorporating conflict gradient surgery named CGS,
% Instead of directly using the adversarial generator which is under the current iteration, we focus on stably improving the robustness of the model by introducing multiple adversarial perturbations.
aiming to 
% always keep the gradient angle $\phi_{ij} \textgreater 0$ by progressively integrating the adversarial data, 
change the distribution of $cos(\theta)$ tends towards positive values, where $\theta$ is the angle between the gradient of the original data and the gradient of the CGS adversarial data, so that the gradient update guided by the adversarial data points in the direction of the conflict-free direction, and finally improves the robustness of the model to perturbations:
% \vspace*{-3mm}
\begin{equation*}
% \vspace*{-3mm}
\mathbb{E}\left[\cos(\theta)\right] > 0, \quad \text{where } \cos(\theta_i) = \frac{g_i \cdot g_{\text{mix}_i}}{|g_i| |g_{\text{mix}_i}|}
% \vspace*{-1.5mm}
\end{equation*}
% \vspace*{-0.5mm}
where $g_i$ is the gradient of the i-th original data, $g_{mix_i}$ is the gradient of the i-th mixed adversarial data. 
We calculate the gradient cosine similarity between the gradient of the adversarial-perturbed data and the original data.  We define two gradients as conflicting if their cosine similarity is negative, indicating they are diverging from one another.  If the distribution of cosine similarities tends to be negative, it signifies the presence of numerous adversely oriented gradients, which can mislead the gradient optimization process. As illustrated in Fig.~\ref{fig:grad_dist}, we extracted 1000 images from the KITTI dataset and recorded the gradient cosine similarity between pairs of adversarial perturbation data and original data. Without employing conflict gradient surgery, the gradient conflict becomes a prevalent issue as the adversarial training progresses iteratively, consequently leading slower convergence and performance degradation.
% \vspace*{-5mm}

Experimental observations provided in Fig.~\ref{fig:grad_dist} demonstrates that our approach ensures the expectation of $cos(\theta)$ is positive,  thereby skewing the distribution of $cos(\theta)$ tends towards positive values,
so that the gradient update guided by the adversarial data points in the direction of the conflict-free direction, ultimately enhancing the model's robustness against perturbations.  
% The expectation of $cos(\theta)$ is greater than 0, indicating that the distribution of $cos(\theta)$ tends towards positive values,
% so that the gradient update guided by the adversarial data points in the direction of the conflict-free direction, and finally improves the robustness of the model to perturbations.  
Specifically, given a set of gradient vectors (one for each adversarial data augmentation), we construct the adversarial perturbation data while employing multiple adversarial generators from previous iterations, restraining the over-regularization caused by the conflict component through our conflict gradient surgery. The update procedure of the CGS algorithm is provided in Algorithm~\ref{alg:scat}.

\begin{algorithm}[t]
\scriptsize
\caption{Generic SCAT self-supervised MDE algorithm
\\ \makebox[\linewidth][r]{(\RedTriangle\ vanilla self-supervised MDE, \BlueTriangleRight\ our modifications)}}
\label{alg:scat}
\begin{algorithmic}[1]
% \State $\theta, \theta_T, \psi$: randomly initialized network parameters, $\psi \leftarrow \theta$ \hfill \BlueTriangleRight {Initialize $\psi$ to be equal to $\theta$}
% \State $\eta, \zeta$: learning rate and momentum coefficient
% \State $\alpha, \beta$: loss coefficients, \textit{default}: ($\alpha = 0.5, \beta = 0.5$)

\For{Epoch $e = 1...n$ do}
\State \textbf{Adversarial Data Augmentation via CGS:}
\State \hspace{1em} $g_{1:j}\sim Sample(\mathcal{B}, j) $ \hfill\BlueTriangleRight {Randomly select generators from History Buffer.}
\State \hspace{1em} $\delta^{1:j} = g_{1:j} (I_t)$ \hfill \BlueTriangleRight {Generate adversarial noise from clean images.}

\State \hspace{1em} $\tilde{I}_{t^{'}}^{1:j} = I_{t} + \delta^{1:j}$ \hfill \BlueTriangleRight {Obtain adversarial images.}

\State \textbf{Deterministic Reprojection with SDN:}
\State \hspace{1em} $D_{t}=ScalingDepthNet(I_t), \; P_{t}=PoseNet(I_t)$ 

\RedTriangle \;{Obtain clean depth map and camera poses.
% via improved MDE with \textbf{SDN}
}
\State \hspace{1em} $\tilde{D}_{t}=ScalingDepthNet(\tilde{I}_t), \; \tilde{P}_{t}=PoseNet(\tilde{I}_t)$ 

\BlueTriangleRight \;{Obtain adversarial depth map and camera poses.
% as the same process.
}

% \State \hspace{1em} $\mathcal{B} \leftarrow \mathcal{B} \cup G_e$ 
\State \hspace{1em} $\tilde{I}_{t^{\prime}\to t} \Leftarrow f_{\theta}(\tilde{I}_{t'}), \; I_{t^{\prime}\to t} \Leftarrow f_{\theta}(I_{t'})$

\State \hspace{1em} $I_t \simeq  \tilde{I}_{t^{\prime}\to t}, \; I_{t^{\prime}\to t} $

\BlueTriangleRight \; {Reconstruct adversarial and clean images with the same target.}
\State \hspace{1em} $\mathcal{B} \leftarrow \mathcal{B} \cup g_e$ \hfill \BlueTriangleRight {Add current Generator into History Buffer $\mathcal{B}$.}
\State \textbf{Module Optimization via SGD:}
\State \hspace{1em} $\theta \leftarrow \theta - \eta_{\theta} \nabla\theta \mathcal{L}_P (f, I_t; \theta)$ \hfill \RedTriangle {Optimize MDE $f_\theta$ with $\mathcal{L}_{P}$ in Eq.~\ref{eq_p}.}
\State \hspace{1em} $\phi \leftarrow \phi + \eta_{\phi} \nabla\phi \mathcal{L}_{AD} (g_{\phi}, I_t; \phi)$ \hfill \BlueTriangleRight {Optimize Generator $g_{\phi}$ with $\mathcal{L}_{AD}$in Eq.~\ref{eq_ad}.}
\EndFor
\end{algorithmic}
\end{algorithm}
\section{Experiments}
%In this section, we adopt KITTI~\cite{geiger2012kitti} as our training dataset and validate the generalization performance on several challenging cross-domain datasets, we also validate the stability enhancement on multiple state-of-the-art models. In addition, we analyze the disadvantages of existing offline data augmentation methods in terms of training efficiency, where our method does not require additional data preparation and can converge quickly. Further, we analyze the impact of high-variance reconstruction targets on self-supervised MDE methods. Finally, we perform an ablation analysis of our SCAT method, verifying the effectiveness of each component for improving generalization capability. For more results about parameter ablations and training efficiency, please refer to the Appendix.
\begin{table*}  % 使用table*创建双栏宽度的表格
    \vspace{-8pt}
    \scriptsize
     \caption{\textbf{Quantitative Results for the KITTI-C Dataset.} \colorbox{best}{Pink} cell denotes the best performance, while \colorbox{third}{yellow} one denotes suboptimal.
     % We evaluated our method on KITTI-C across a total of 18 damage types, with each type comprising 5 different damage levels for qualitative comparison. The results demonstrate that our approach significantly improves robustness metrics, such as $mCE$ (mean corruption error), and depth estimation metrics like $Abs\;Rel$ (absolute relative error), establishing SOTA performance on the KITTI-C dataset. Note that \textbf{bold} text represents the best results for each test metric, while text with an \underline{underline} represents the second-best results. Additionally, green metrics indicate better performance as the values decrease, while pink indicates the opposite.
     }
    % \vspace{-8pt}
    \centering
    \renewcommand{\arraystretch}{1.2}
    \begin{adjustbox}{width=\textwidth,center}
    \begin{tabular}{l|c|ccccccccc}
        \hline
       % Method & Data & W $\times$ H &  \cellcolor[RGB]{127,255,212} mCE \newline ($\%$) &  \cellcolor[RGB]{255,182,193} mRR \newline ($\%$) & \cellcolor[RGB]{127,255,212} Abs Rel & \cellcolor[RGB]{127,255,212} Sq Rel & \cellcolor[RGB]{127,255,212} RMSE & \cellcolor[RGB]{127,255,212} RMSE log & \cellcolor[RGB]{255,182,193} $\delta<1.25$ & \cellcolor[RGB]{255,182,193} $\delta<1.25^2$ & \cellcolor[RGB]{255,182,193} $\delta<1.25^3$\\
       Method & W $\times$ H &  mCE \newline ($\%$)$\downarrow$ &   mRR \newline ($\%$)$\uparrow$ &  Abs Rel$\downarrow$ & Sq Rel$\downarrow$ &  RMSE$\downarrow$ &  RMSE log$\downarrow$ & $\delta<1.25\uparrow$ &  $\delta<1.25^2\uparrow$ & $\delta<1.25^3\uparrow$\\
        \hline
        MonoDepth2~\cite{godard2019monodepth2} & $640 \times 192$ & 101.04 & 84.08 & 0.248 & 1.764 & 6.852 & 0.291 & 0.698 & 0.874 & 0.944\\
        + DA &  $640 \times 192$ & 97.72 & 84.96 & 0.227 & 1.761 & 6.735 & 0.293 & 0.710 & 0.876 & 0.944 \\
        + ADA & $640 \times 192$ & \cellcolor{third}89.31 & \cellcolor{third}87.31 & \cellcolor{third}0.193 & \cellcolor{third}1.632 & \cellcolor{third}6.427 & \cellcolor{third}0.278 & \cellcolor{third}0.734 & \cellcolor{third} 0.882 & \cellcolor{third}0.945 \\
        + \textbf{SCAT\;(ours)}  & $640 \times 192$ & \cellcolor{best}\textbf{86.32} & \cellcolor{best}\textbf{90.13} & \cellcolor{best}\textbf{0.165} & \cellcolor{best}\textbf{1.573} & \cellcolor{best}\textbf{6.157} & \cellcolor{best}\textbf{0.269} & \cellcolor{best}\textbf{0.762} & \cellcolor{best}\textbf{0.893} & \cellcolor{best}\textbf{0.951}\\
        \hline
        CADepth~\cite{yan2021cadepth}  & $640 \times 192$ & 108.10 & 80.12 & 0.271 & 1.902 & 6.900 & 0.302 & 0.674 & 0.860 & 0.938 \\
        + DA  &  $640 \times 192$ & 107.14 & 81.31 & 0.261 & 1.874 & 6.832 & 0.299 & 0.691 & 0.865 & 0.939 \\
        + ADA & $640 \times 192$ & \cellcolor{third}99.81 & \cellcolor{third}83.13 & \cellcolor{third}0.235 & \cellcolor{third}1.866 & \cellcolor{third}6.565 & \cellcolor{third}0.278 & \cellcolor{third}0.721 & \cellcolor{third}0.871 & \cellcolor{third}0.940 \\
        + \textbf{SCAT\;(ours)}  &  $640 \times 192$ & \cellcolor{best}\textbf{95.54} & \cellcolor{best}\textbf{87.93} & \cellcolor{best}\textbf{0.197} & \cellcolor{best}\textbf{1.794} & \cellcolor{best}\textbf{6.258} & \cellcolor{best}\textbf{0.264} & \cellcolor{best}\textbf{0.747} & \cellcolor{best}\textbf{0.883} & \cellcolor{best}\textbf{0.948} \\
        \hline
        HR-Depth~\cite{lyu2021hrdepth}  & $640 \times 192$ & 103.67 & 82.88 & 0.255 & 1.960 & 6.909 & 0.309 & 0.670 & 0.854 & 0.933 \\
        + DA  &  $640 \times 192$ & 101.74 & 83.23 & 0.243 & 1.942 & 6.787 & 0.294 & 0.685 & 0.859 & 0.936\\
        + ADA &  $640 \times 192$ & \cellcolor{third}93.84 & \cellcolor{third}85.32 & \cellcolor{third}0.215 & \cellcolor{third}1.831 & \cellcolor{third}6.625 & \cellcolor{third}0.278 & \cellcolor{third}0.699 & \cellcolor{third}0.861 & \cellcolor{third}0.939 \\
        + \textbf{SCAT\;(ours)}   & $640 \times 192$ & \cellcolor{best}\textbf{87.06} & \cellcolor{best}\textbf{89.47} & \cellcolor{best}\textbf{0.189} & \cellcolor{best}\textbf{1.732} & \cellcolor{best}\textbf{6.204} & \cellcolor{best}\textbf{0.258} & \cellcolor{best}\textbf{0.741} & \cellcolor{best}\textbf{0.871} & \cellcolor{best}\textbf{0.945} \\
        \hline
        MonoVit~\cite{zhao2021m& MonoVit}  & $640 \times 192$ & 80.54 & 88.98 & 0.191 & 1.274 & 5.998 & 0.245 & 0.771 & 0.922 & 0.964 \\
        + DA  & $640 \times 192$ & 78.32 & 90.47 & 0.187 & 1.256 & 5.873 & 0.228 & 0.787 & 0.931 & 0.965 \\
        + ADA  & $640 \times 192$ & \cellcolor{third}72.41 & \cellcolor{third}92.34 & \cellcolor{third}0.166 & \cellcolor{third}1.134 & \cellcolor{third}5.626 & \cellcolor{third}0.211 & \cellcolor{third}0.819 & \cellcolor{third}0.935 & \cellcolor{third}0.969 \\
        + \textbf{SCAT\;(ours)}  & $640 \times 192$ & \cellcolor{best}\textbf{62.74} & \cellcolor{best}\textbf{95.38} & \cellcolor{best}\textbf{0.127} & \cellcolor{best}\textbf{1.058} & \cellcolor{best}\textbf{5.274} & \cellcolor{best}\textbf{0.208} & \cellcolor{best}\textbf{0.846} & \cellcolor{best}\textbf{0.947} & \cellcolor{best}\textbf{0.976} \\
        \hline
        Robust-Depth*~\cite{saunders2023self}   & $640 \times 192$ & 55.72 & 96.46 & 0.121 & 0.954 & 5.051 & 0.201 & 0.854 & 0.952 & 0.978 \\
        + DA   & $640 \times 192$ & 60.12 & 96.34 & 0.124 & 0.963 & 5.012 & 0.213 & 0.852 & 0.951 &  0.977 \\
        + ADA  & $640 \times 192$ & \cellcolor{third}54.23 & \cellcolor{third}97.59 & \cellcolor{third}0.119 & \cellcolor{third}0.943 & \cellcolor{third}4.985 & \cellcolor{third}0.199 & \cellcolor{third}0.853 & \cellcolor{third}0.951 & \cellcolor{third}0.977 \\
        + \textbf{SCAT\;(ours)}   & $640 \times 192$ & \cellcolor{best}\textbf{53.37} & \cellcolor{best}\textbf{98.19} & \cellcolor{best}\textbf{0.117} & \cellcolor{best}\textbf{0.932} & \cellcolor{best}\textbf{4.872} & \cellcolor{best}\textbf{0.187} & \cellcolor{best}\textbf{0.861} & \cellcolor{best}\textbf{0.955} & \cellcolor{best}\textbf{0.979} \\
        \hline
    \end{tabular}
    \end{adjustbox}
    \label{tab:kittic_res}
\vspace{-15pt}
\end{table*}

\subsection{Experimental Settings}
\noindent\textbf{KITTI~\cite{geiger2012kitti}} 
KITTI dataset is collected for mobile robotics and autonomous driving, involving hours of traffic scenarios recorded with a variety of sensor modalities, including high-resolution RGB, grayscale stereo cameras, and a 3D laser scanner. 
Following the setting in~\cite{zhou2017sfm}, we use 39,810 images for training and 4,424 for validation. Subsequently, we rigorously evaluate the proposed method and other compared methods on the KITTI eigen test dataset~\cite{eigen2014depth}.

\noindent\textbf{KITTI-C~\cite{kong2023robodepth}} To evaluate the robustness and safety of our method in out-of-distribution (OoD) scenarios, we utilize KITTI-C dataset as a benchmark.
% KITTI-C involves 18 common corruption patterns, spanning variations in weather and lighting conditions, sensor failures, movement, and noises during data processing. 
The diverse corruptions enable the reliability of simulation to the potential perturbation distribution in real-world scenarios. 

\noindent\textbf{DrivingStereo~\cite{wang2018drivingstereo}} It contains 500 images captured under different weather conditions, involving foggy, cloudy, rainy, and sunny. These images offer a realistic representation of various driving scenarios, enabling robust evaluation of depth estimation performance across diverse domains.

\noindent\textbf{Foggy CityScapes~\cite{sakaridis2018semantic}} Foggy CityScapes is a synthetic fog dataset that simulates real-world foggy scenarios. Each foggy sample is generated using the corresponding clear image and depth map from CityScapes dataset.

\noindent\textbf{NuScenes~\cite{nuscenes}} The NuScenes dataset comprises approximately 15,000 images captured in real nighttime urban street settings. To validate the model generalization, NuScenes is also selected as an evaluation benchmark in our study. 

\begin{table*}
    \scriptsize
    \vspace{-15pt}
     \caption{%Comparison results in terms of Abs Rel, Sq Rel, RMSE and RMSE log on the KITTI Dataset with the image size of $640\times 192$.
    \textbf{Quantitative Results for the KITTI Eigen Test Dataset.}
    % To validate the depth estimation performance of our framework on clean datasets, we conducted a quantitative comparison on the KITTI eigen test dataset.
    The results indicate that SCAT largely maintains the performance of all baselines compared to vanilla ADA, ensuring the ability to infer accurate depth information on clean images.
    }
    \centering
    \renewcommand{\arraystretch}{1.2}
    \begin{adjustbox}{width=\textwidth,center}
    \begin{tabular}{l|c|ccccccc}
        \hline
        Method & W $\times$ H & Abs Rel$\downarrow$ & Sq Rel$\downarrow$ & RMSE$\downarrow$ & RMSE log$\downarrow$ & $\delta<1.25\uparrow$ &  $\delta<1.25^2\uparrow$ & $\delta<1.25^3\uparrow $ \\
        \hline
        Monodepth2~\cite{godard2019monodepth2}  & $640 \times 192$ &  \cellcolor{best}0.115 &  \cellcolor{best}0.903 & \cellcolor{best}4.863 & \cellcolor{best}0.193 & \cellcolor{best}0.877 & \cellcolor{best}0.959 &  \cellcolor{best}0.981 \\
        % + DA   & $640 \times 192$  &  &  &  &  &  &  &  \\
        + ADA  & $640 \times 192$  & 0.121 & 0.978 & 4.992 & 0.221 & 0.862 & 0.953 & 0.978 \\
        + \textbf{SCAT\;(ours)}   & $640 \times 192$  & \cellcolor{third}0.116 & \cellcolor{third}0.942 & \cellcolor{third}4.877 & \cellcolor{third}0.193 & \cellcolor{third}0.877 & \cellcolor{third}0.958  & \cellcolor{third}0.981 \\
        \hline
        CADepth~\cite{yan2021cadepth}  & $640 \times 192$ &  \cellcolor{best}0.105 &  \cellcolor{best}0.769 &  \cellcolor{best}4.535 &  \cellcolor{best}0.181 & \cellcolor{best}0.892 & \cellcolor{best}0.964 &  \cellcolor{best}0.983 \\
        % + DA   & $640 \times 192$  &  &  &  &  &  &  &  \\
        + ADA  & $640 \times 192$  & 0.108 & 0.812 & 4.585 & 0.190 & 0.887 & 0.962 & 0.982 \\
        + \textbf{SCAT\;(ours)}   & $640 \times 192$ & \cellcolor{third}0.106 & \cellcolor{third}0.796 & \cellcolor{third}4.579 & \cellcolor{third}0.183 & \cellcolor{third}0.890 & \cellcolor{third}0.963 & \cellcolor{third}0.983 \\
        \hline
        DIFFNet~\cite{zhou2021diffnet}  & $640 \times 192$ & \cellcolor{best}0.102 &  \cellcolor{best}0.749 & \cellcolor{best}4.445 & \cellcolor{best}0.179 & \cellcolor{best}0.897 & \cellcolor{best}0.965 & \cellcolor{best}0.983 \\
        % + DA   & $640 \times 192$ &  &  &  &  &  &  &  \\
        + ADA  & $640 \times 192$ & 0.106 & 0.793 & 4.592 & 0.184 & 0.889 & 0.963 & 0.982 \\
        + \textbf{SCAT\;(ours)}   & $640 \times 192$ & \cellcolor{third}0.103 & \cellcolor{third}0.764 & \cellcolor{third}4.453 & \cellcolor{third}0.180 & \cellcolor{third}0.897 & \cellcolor{third}0.965 & \cellcolor{third}0.983 \\
        \hline
        MonoVit~\cite{zhao2021m& MonoVit}  & $640 \times 192$ &  \cellcolor{best}0.099 & \cellcolor{best}0.708 &  \cellcolor{best}4.372 & \cellcolor{best}0.175 & \cellcolor{best}0.900 & \cellcolor{best}0.967 & \cellcolor{best}0.984\\
        % + DA   & $640 \times 192$   &  &  &  &  &  &  &  \\
        + ADA  & $640 \times 192$  & 0.106 & 0.733 & 4.591 & 0.184 & 0.897 & 0.963 & 0.982 \\
        + \textbf{SCAT\;(ours)}   & $640 \times 192$  & \cellcolor{third}0.100 & \cellcolor{third}0.716 & \cellcolor{third}4.389 & \cellcolor{third}0.175 & \cellcolor{third}0.899 & \cellcolor{third}0.967 & \cellcolor{third}0.984 \\
        \hline
        Robust-Depth*~\cite{saunders2023self}  & $640 \times 192$ &  \cellcolor{best}0.100 &  \cellcolor{best}0.747 & \cellcolor{best}4.455 &  \cellcolor{best}0.177 & \cellcolor{best}0.895 & \cellcolor{best}0.966 & \cellcolor{best}0.984 \\
        % + DA   & $640 \times 192$ &  &  &  &  &  &  &  \\
        + ADA  & $640 \times 192$ & 0.105 & 0.783 & 4.698 &0.182 & 0.889 & 0.953 & 0.981\\
        + \textbf{SCAT\;(ours)}   & $640 \times 192$ & \cellcolor{third}0.101 & \cellcolor{third}0.754 & \cellcolor{third}4.459 & \cellcolor{third}0.177 & \cellcolor{third}0.895 & \cellcolor{third}0.966 & \cellcolor{third}0.984 \\
        \hline
    \end{tabular}
    \end{adjustbox}
    \label{tab:kitti_res}
\vspace{-20pt}
\end{table*}

\subsection{Effect of Perturbation Sizes}
\begin{figure}[t]
%\vspace{-10pt}
\centering
\includegraphics[width=\columnwidth, keepaspectratio, interpolate=true]{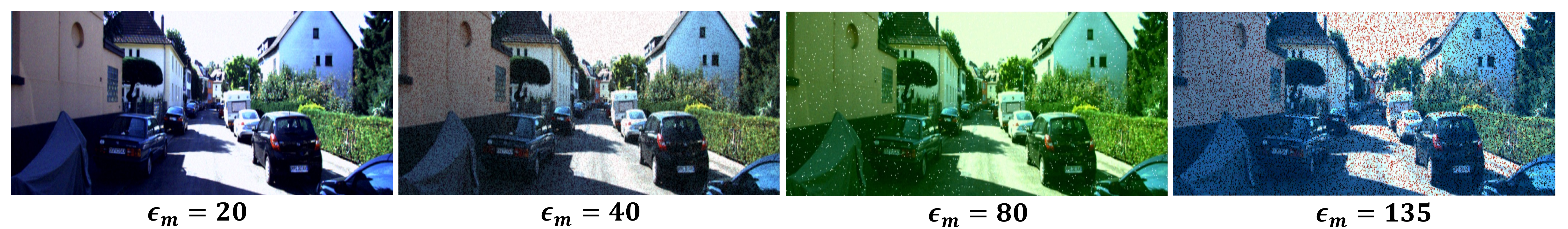}
\vspace{-20pt}
\caption{Example images with different $\epsilon_m$ of adversarial perturbation.} 
% on MonoDepth2.}
\vspace{-15pt}
\label{fig:epsilon}
\end{figure}

To evaluate the effect of adversarial perturbation magnitude on model generalization, here we employ $\epsilon_m$ to perturbation sizes:
\begin{equation}
\label{eq_epsilon}
\epsilon_m = \min \lVert \delta_{\min} \rVert_{2}.
\end{equation}
We measure model generalization capability by calculating the median perturbation size $\epsilon_m$ and report the results in Tab.~\ref{tab:noisy_size}. To provide a better intuition for the noise level for particular $\epsilon_m$, we display example images in Fig.~\ref{fig:epsilon}.

\begin{table*} [t] % 使用table*创建双栏宽度的表格
    %\vspace{-30pt}
    % \small
    \scriptsize
     \caption{\textbf{Qualitative Results for different perturbation sizes.} We compare the results obtained by $\epsilon_m=135.0$ adversarial noise generator with its counterparts of different perturbation sizes. Note that $\epsilon_m=135.0$ was used in all experiments in this paper apart from this ablation study.
     }
    % \vspace{-3pt}
    \centering
    \renewcommand{\arraystretch}{1.2}
    \begin{adjustbox}{width=\textwidth,center}
    \begin{tabular}{l|c|c|ccccccccc}
    \hline
    \textbf{Model} & MCE \newline ($\%$)$\downarrow$ &   mRR \newline ($\%$)$\uparrow$ &  Abs Rel$\downarrow$ & Sq Rel$\downarrow$ &  RMSE$\downarrow$ &  RMSE log$\downarrow$ & $\delta<1.25\uparrow$ &  $\delta<1.25^2\uparrow$ & $\delta<1.25^3\uparrow$\\
    \hline
    MonoDepth2~\cite{godard2019monodepth2} &  101.04 & 84.08 & 0.248 & 1.764 & 6.852 & 0.291 & 0.698 & 0.874 & 0.944 \\
    \textbf{+ $g_{\phi}\; \epsilon_m = 20$} &  93.96 & 86.17 & 0.196 & 1.619 & 6.385 & 0.285 & 0.738 & 0.883 & 0.947 \\
     \textbf{+ $g_{\phi}\; \epsilon_m = 40$}   &  90.15 & 86.93 & 0.187 & 1.613 & 6.351 & 0.282 & 0.741 & 0.885 & 0.949 \\
     \textbf{+ $g_{\phi}\; \epsilon_m = 80$}  & 88.95 & \cellcolor{third}87.21 & 0.182 & 1.602 & 6.288 & 0.279 & 0.749 & 0.890 & 0.949 \\
     \textbf{+ $g_{\phi}\; \epsilon_m = 135$}  & \cellcolor{best}\textbf{86.32} & \cellcolor{best}\textbf{90.13} & \cellcolor{best}\textbf{0.165} & \cellcolor{best}\textbf{1.573} & \cellcolor{best}\textbf{6.157} & \cellcolor{best}\textbf{0.269} & \cellcolor{best}\textbf{0.762} & \cellcolor{best}\textbf{0.893} & \cellcolor{best}\textbf{0.951} \\
     \textbf{+ $g_{\phi}\; \epsilon_m = 180$}  & \cellcolor{third}88.17 & 86.83 & \cellcolor{third}0.177 & \cellcolor{third}1.594 & \cellcolor{third}6.195 & \cellcolor{third}0.277 & \cellcolor{third}0.758 & \cellcolor{third}0.891 & \cellcolor{third}0.950 \\
    \hline
    \end{tabular}
    \end{adjustbox}
    \label{tab:noisy_size}
\vspace{-2pt}
\end{table*}

\begin{figure}[t]
\centering
\includegraphics[width=\columnwidth, keepaspectratio, interpolate=true]{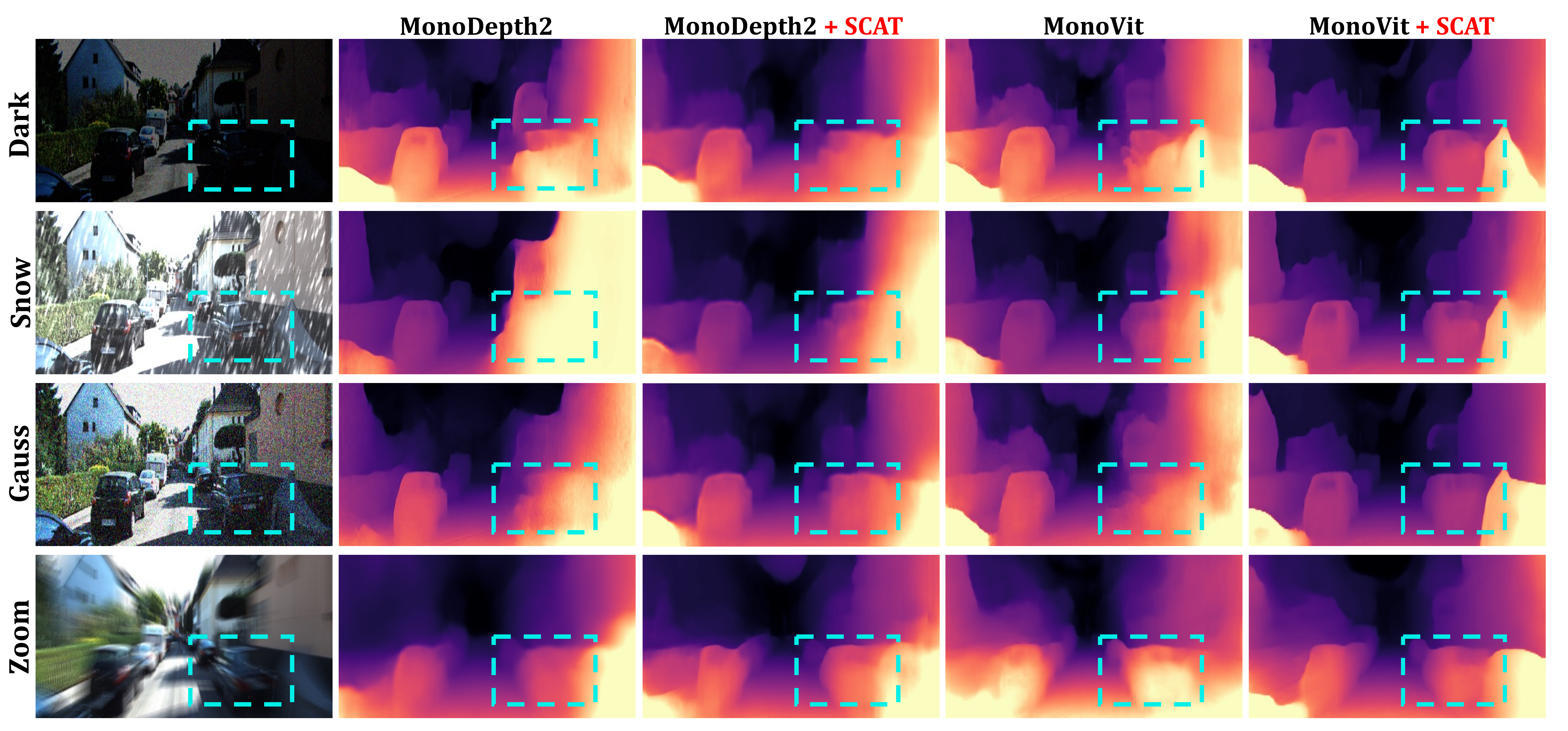}
\caption{\textbf{Qualitative Results for KITTI-C.} As the SOTA self-supervised MDE methods, MonoDepth and MonoVit excel on the KITTI dataset, but struggle to accurately infer depth information from various types of damaged images in out-of-distribution (OoD) domains. With our SCAT framework, their depth estimation performance in challenging cross-domain scenarios can be substantially improved.}%%%%\vspace*{-6mm}
\label{fig:SCAT_kittic}
\vspace{-10pt}
\end{figure}
\begin{figure}[t]
\centering
\includegraphics[width=\columnwidth, keepaspectratio, interpolate=true]{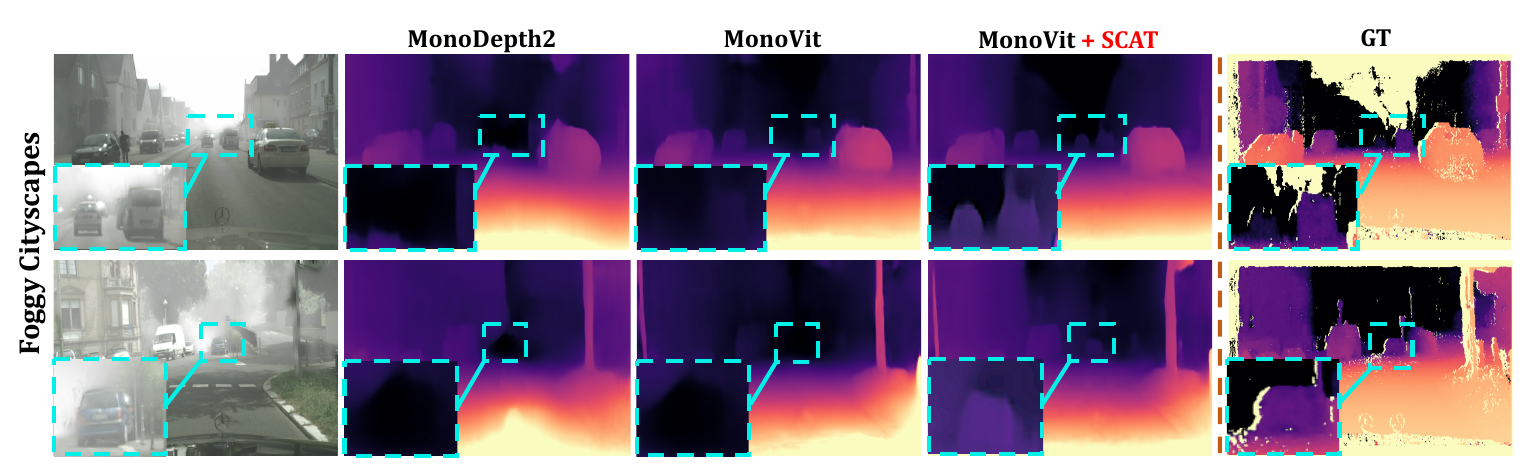}
\caption{\textbf{Qualitative Results for Foggy CityScapes.} We conduct testing to assess SCAT's generalization on a synthetically generated dataset with more severe fog. 
% In comparison to vanilla methods, SCAT-based models exhibit higher accuracy in predicting the depth of distant objects obscured by the fog.
}
% %%%%\vspace*{-3mm}
\label{fig:foggycity}
\end{figure}
\subsection{Results on KITTI-C and KITTI}
In this section, we compare the depth estimation performance of our SCAT  on the KITTI-C and KITTI datasets. Quantitative results on the KITTI-C dataset are shown in Tab.~{\ref{tab:kittic_res}}. We conducted extensive experiments on multiple baseline models,  comparing it with the generalization performance between offline data augmentation (DA) and vanilla adversarial data augmentation (ADA). 
As expected, our method demonstrates a significant advantage on the mean corruption error (mCE) and mean resilience rate (mRR) scores. Especially for 18 common out-of-distribution corruption types, with all baseline models, our proposed SCAT method outperforms the competitors in almost all of metrics, providing a general framework to enhance cross-domain generalization capability for existing self-supervised monocular depth estimation methods.

\begin{table}[h]
    \scriptsize
    % \caption{\textbf{Quantitative Results for the Real-World NuScenes-Night and Foggy CityScapes Dataset.}}
    \caption{\textbf{Quantitative Results for the Real-World NuScenes-Night Dataset.}}
    % \vspace{-3pt}
    \centering
    \renewcommand{\arraystretch}{1.2}
    \begin{adjustbox}{width=\textwidth,center}
    \begin{tabular}{l|l|ccccccc}
        \hline
        Dataset & Method & Abs Rel$\downarrow$ & Sq Rel$\downarrow$ &  RMSE$\downarrow$ &  RMSE log$\downarrow$ & $\delta<1.25\uparrow$ & $\delta$$<1.25^2\uparrow$ & $\delta<1.25^3\uparrow$ \\
        \hline
        \multirow{10}{*}{\parbox{3.1cm}{NuScenes-Night~\cite{nuscenes}}} & Monodepth2~\cite{godard2019monodepth2} & 0.397 & 6.206 & 14.569 & 0.568 & 0.378 & 0.650 & 0.794 \\
        & + \textbf{SCAT\;(ours)}  & \cellcolor{best}\textbf{0.355} & \cellcolor{best}\textbf{5.112} & \cellcolor{best}\textbf{12.436} & \cellcolor{best}\textbf{0.499} & \cellcolor{best}\textbf{0.392} & \cellcolor{best}\textbf{0.671} & \cellcolor{best}\textbf{0.863} \\
        \cline{2-9}
        & HR-Depth~\cite{lyu2021hrdepth} & 0.461&  6.633& 15.028& 0.622& 0.301& 0.571& 0.749 \\
        & + \textbf{SCAT\;(ours)}  & \cellcolor{best}\textbf{0.424} & \cellcolor{best}\textbf{5.802} & \cellcolor{best}\textbf{13.685} & \cellcolor{best}\textbf{0.602} & \cellcolor{best}\textbf{0.324} & \cellcolor{best}\textbf{0.584} & \cellcolor{best}\textbf{0.827} \\
        \cline{2-9}
        & CADepth~\cite{yan2021cadepth} & 0.421 &5.949& 14.509& 0.593& 0.331& 0.613& 0.776 \\
        & + \textbf{SCAT\;(ours)}  & \cellcolor{best}\textbf{0.387} & \cellcolor{best}\textbf{5.524} & \cellcolor{best}\textbf{11.962} & \cellcolor{best}\textbf{0.577} & \cellcolor{best}\textbf{0.365} & \cellcolor{best}\textbf{0.655} & \cellcolor{best}\textbf{0.834} \\
        \cline{2-9}
        & MonoVit~\cite{zhao2021m& MonoVit} & 0.313 & 4.143 & 12.252& 0.455& 0.485& 0.736& 0.858\\
        & + \textbf{SCAT\;(ours)}  & \cellcolor{best}\textbf{0.284} & \cellcolor{best}\textbf{4.126} & \cellcolor{best}\textbf{10.139} & \cellcolor{best}\textbf{0.401} & \cellcolor{best}\textbf{0.497} & \cellcolor{best}\textbf{0.752} & \cellcolor{best}\textbf{0.879} \\
        \cline{2-9}
        & Robust-Depth*~\cite{saunders2023self} & 0.276& 4.075& 10.470& 0.380& 0.607& 0.819 & 0.912 \\
        & + \textbf{SCAT\;(ours)}  & \cellcolor{best}\textbf{0.263} & \cellcolor{best}\textbf{3.973} & \cellcolor{best}\textbf{9.462} & \cellcolor{best}\textbf{0.371} & \cellcolor{best}\textbf{0.609} & \cellcolor{best}\textbf{0.819} & \cellcolor{best}\textbf{0.913} \\
        \hline
    \end{tabular}
    \label{tab:Nuscenes_night}
    \end{adjustbox}
\vspace{-10pt}
\end{table}

\subsection{Results on Cross-Domain Generalization}
\begin{figure*}[t]
\vspace{-5pt}
\centering
\includegraphics[width=\textwidth, keepaspectratio, interpolate=true]{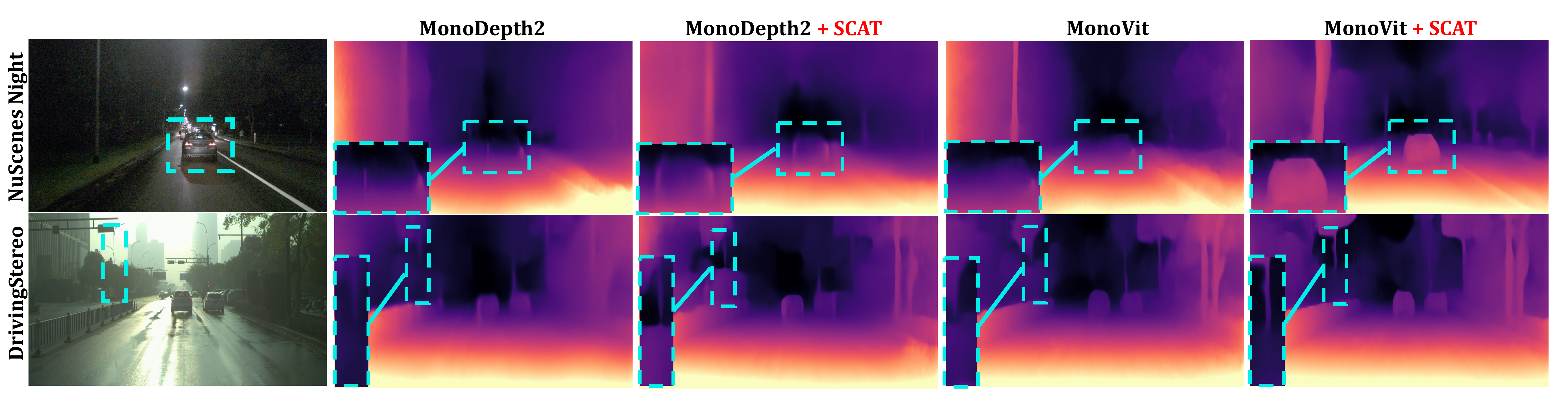} % Reduce the figure size so that it is slightly narrower than the column. Don't use precise values for figure width.This setup will avoid overfull boxes.
\caption{\textbf{Qualitative results on Driving Stereo and Nuscenes.} %To validate the generalization performance of ARDepth on real cross-domain datasets, we conducted a qualitative analysis on the nighttime scenes of the NuScenes dataset and the DrivingStereo dataset with complex weather conditions. 
It can be observed that our SCAT method effectively recovers challenging distant objects such as vehicles or streetlights when dealing with complex and unknown images in real scenarios.}%%%%\vspace*{-3mm}
\label{fig:real-wrold}
\vspace{-20pt}
\end{figure*}

To demonstrate the generalization capability of our SCAT %of this method
over real cross-domain scenarios, like foggy and nighttime samples, we also
conduct the experiments 
% with other competitors %in depth estimation performance 
on Foggy CityScapes dataset, DrivingStereo dataset and NuScenes dataset, which are shown in Fig.~\ref{fig:foggycity} and Fig.~\ref{fig:real-wrold}. %nighttime scenes from the NuScenes dataset. The q
As expected, our SCAT generates depth estimation results with more reliable and clearer contours in all weather conditions. By contrast, vanilla MonoDepth2 and MonoVit suffer from significant performance decline when handling real and unknown scenarios, in particular in the rainy and nighttime scenes. 
Notably, it can be observed that our SCAT method can effectively recover challenging distant objects such as vehicles or streetlights when dealing with complex and unknown images in real-world cross-domain scenarios.
Quantitative results are summarized in Tab.~{\ref{tab:Nuscenes_night}}, showing that our proposed SCAT framework gains better performance on all metrics with multiple state-of-the-art methods. More results are shown in the Appendix.
% Visual comparisons on real-world scenarios, including Driving Stereo and Nuscenes datasets are shown in Fig.~\ref{fig:real-wrold}. We also provide the cross-domain comparison on Foggy CityScapes in Fig.~\ref{fig:foggycity}. 

\subsection{Ablation Study}
\subsubsection{Effectiveness of Individual Components.}
To improve the stability and generalization capability of self-supervised MDE methods, SCAT presents a stabilized adversarial training framework with Conflict Gradient Surgery (CGS), progressively integrating the adversarial-perturbed data to prevent gradient conflict. 
\noindent In addition, we propose a scaling depth network named SDN, reducing the perturbation sensitivity of the UNet-based depth network. We perform an ablation study to investigate the effectiveness of individual components in SCAT and the results are shown in Tab.~\ref{tab:ablation}.  Individually, each of these contributes significantly to the improvement of generalization performance across multiple datasets. 

\vspace{-15pt}
\subsubsection{Impact of LSCs Coefficient $\kappa$.}
In our experiments, the value of the long skip connections (LSCs) coefficient $\kappa$ was set to balance the self-supervised MDE model's stability and generalization capability, which defaults to 0.7 in the paper. To assess the perturbation sensitivity of UNet-based depth networks, we compare the model performance over constant values of $\kappa\in\{0.1, 0.3, 0.7, 1.0\}$. 
Fig.~\ref{fig:fiveline} shows that when $\kappa$ is set to $0.7$, our scaling depth network has extraordinary generalization capability to unseen scenarios in the KITTI-C dataset while keeping the depth estimation performance of the KITTI dataset. 
% 放到补充材料
% When $\kappa$ is set to $0.3$, despite it being proven in Section~\ref{SDN} to result in a smaller reconstruction error and better stability against perturbed images, the long skip connections of the UNet-based depth network are partly limited. This limitation leads to the model's inability to effectively fuse semantic information with spatial information, resulting in a decrease in performance on both KITTI and KITTI-C datasets. Furthermore, when $\kappa$ is set to $0.1$, the LSCs are almost removed, which directly limits the upper bound of its performance.

\begin{minipage}[t]{\linewidth}
\vspace{-20pt}
\begin{minipage}{0.525\linewidth}
% \vspace{-5pt}
\vspace{20pt}  % 调整图片上方的垂直间距
\begin{table}[H]
\renewcommand{\arraystretch}{1.2}
% \footnotesize
\tiny
\centering
\begin{tabularx}{\linewidth}{XX|XXXX}
\hline
\multicolumn{2}{c}{\multirow{2}{*}{}} & \multicolumn{2}{|c}{KITTI} & \multicolumn{2}{c}{KITTI-C} \\
\cline{3-6}
\multicolumn{1}{c}{CGS} & \multicolumn{1}{c|}{SDN} & Abs Rel & $\delta<1.25$ & Abs Rel & $\delta<1.25$ \\
\hline
% \centering CGS & \centering SDN &  &  &  &  \\
 &  & 0.121 & 0.862 & 0.193 & 0.734 \\
 \centering $\checkmark$ & & 0.117 & 0.869 & 0.174 & 0.752 \\
 & \centering $\checkmark$ & 0.118 & 0.865 & 0.179 & 0.748 \\
 \centering $\checkmark$  & \centering $\checkmark$ & 0.116 & 0.877  & 0.165 & 0.762 \\
\hline
\end{tabularx}
%\vspace{5pt}  % 调整图片上方的垂直间距
\caption{Effect of individual components.}
\label{tab:ablation}
\end{table}
\end{minipage}
\hfill
\begin{minipage}{0.45\linewidth}
\centering
\includegraphics[width=0.825\linewidth]{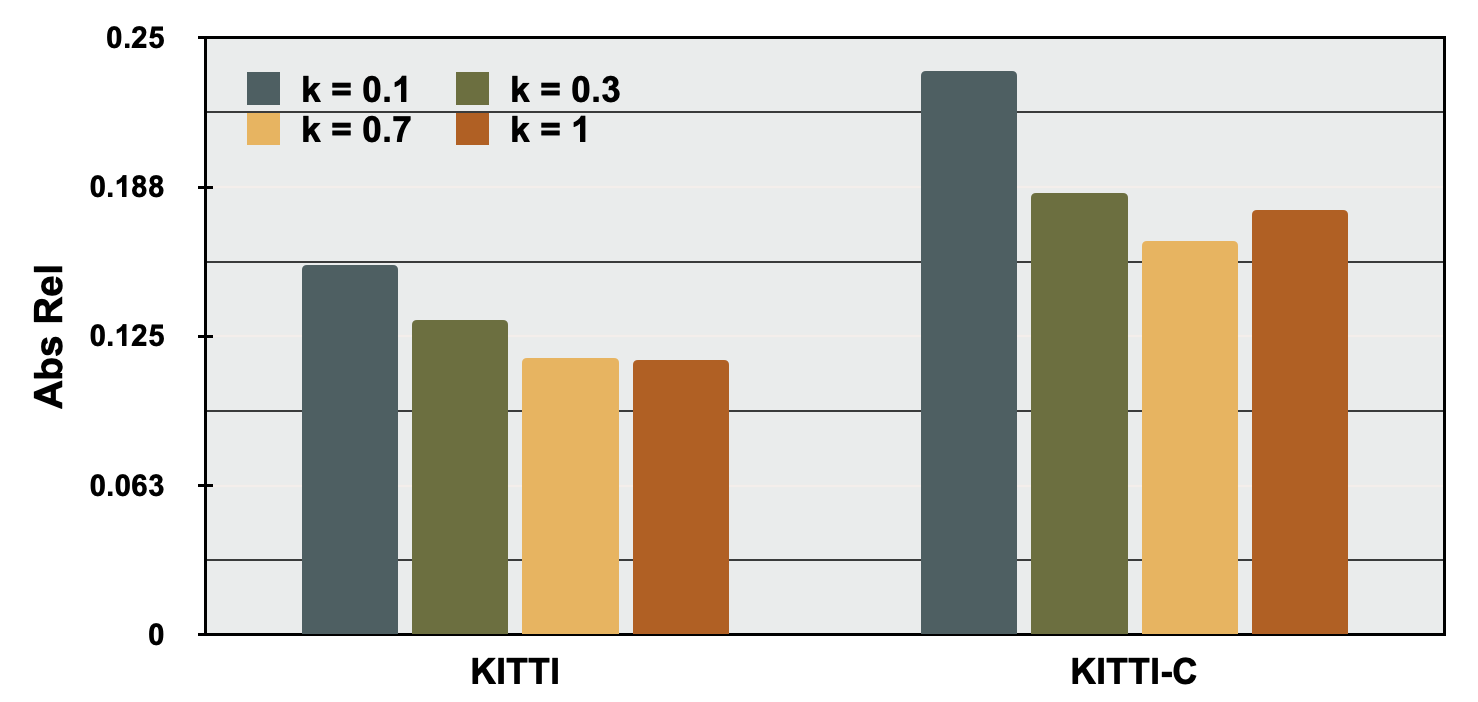}
%\vspace{-1pt}  % 调整图片上方的垂直间距
\captionsetup{aboveskip=-1pt}  % 调整标题上方的间距
\captionof{figure}{Effect of the scaling factor $\kappa$.}
\label{fig:example}
\vspace{-20pt}
\end{minipage}
\vspace{-20pt}
\end{minipage}

\section{Conclusion}
In this work, we initially illuminate the main causes of training collapse under adversarial data augmentation: (i) inherent sensitivity in the UNet-alike depth network and (ii) dual optimization conflict caused by over-regularization. To address these issues, we propose SCAT, a simple yet effective model-agnostic framework that leverages stabilized adversarial training to bootstrap the generalization capability of self-supervised MDE models in unseen scenarios. 
Extensive experiments on a variety of scenarios from five benchmarks validate the merits of our adversarial data augmentation, which endows the model with universal generalization capability and better training efficiency.
In addition, empirical results indicate that each component in our method and the utilization of targeted reconstruction are crucial for generalization performance gain. Our exploration may inspire more researchers to dig into the great potential of utilizing adversarial training in self-supervised MDE.

\section{Limitation}
In the future, we would like to explore iterative adversarial data augmentation~\cite{liu2022eusa}, which is a more promising parameter-efficient solution.

\noindent\textbf{Acknowledgement.} The research was supported by the National Natural Science Foundation of China (U23B2009).

\end{document}